\documentclass{article} 
\usepackage{myStyle_4_DeepDiffEq,times}
\usepackage{amsmath,amsfonts,bm}

\def\eqref#1{equation~\ref{#1}}

\def\1{\bm{1}}

\DeclareMathAlphabet{\mathsfit}{\encodingdefault}{\sfdefault}{m}{sl}
\SetMathAlphabet{\mathsfit}{bold}{\encodingdefault}{\sfdefault}{bx}{n}

\newcommand{\R}{\mathbb{R}}

\usepackage{hyperref,dsfont}
\usepackage{url}
\usepackage{color, colortbl}
\usepackage{amsbsy,amssymb}
\usepackage{booktabs}
\usepackage{verbatim}
\usepackage{graphicx}
\usepackage{wrapfig}
\usepackage{appendix}
\usepackage{multirow,titlesec }
\titlespacing\section{0pt}{4pt plus 1pt minus 1pt}{2pt plus 1pt minus 1pt}
\titlespacing\subsection{0pt}{4pt plus 1pt minus 1pt}{2pt plus 1pt minus 1pt}
\let\OLDitemize\itemize
\renewcommand\itemize{\OLDitemize\addtolength{\itemsep}{0pt}}
\title{Dissipative SymODEN: Encoding Hamiltonian Dynamics with Dissipation and Control into Deep Learning}
\author{Yaofeng Desmond Zhong${^{\dagger}}$, Biswadip Dey${^{\ddagger}}$, and Amit Chakraborty${^{\ddagger}}$ \\
${^{\dagger}}$Princeton University ,   ${^{\ddagger}}$Siemens Corporation, Corporate Technology\\
\texttt{y.zhong@princeton.edu,(biswadip.dey,amit.chakraborty)@siemens.com}}
\iclrfinalcopy 
\begin{document}
%
%
%
\maketitle
%
%
\begin{abstract}
In this work, we introduce \textit{Dissipative SymODEN}, a deep learning architecture which can infer the dynamics of a physical system with dissipation from observed state trajectories. To improve prediction accuracy while reducing network size, Dissipative SymODEN encodes the port-Hamiltonian dynamics with energy dissipation and external input into the design of its computation graph and learns the dynamics in a structured way. The learned model, by revealing key aspects of the system, such as the inertia, dissipation, and potential energy, paves the way for energy-based controllers. 
\end{abstract}
%
%
%

\section{Introduction}
%
Inferring systems dynamics from observed trajectories plays a critical role in identification and control of complex, physical systems, such as robotic manipulators \cite{lillicrap2015continuous} and HVAC systems \cite{Wei:2017:DRL:3061639.3062224}. Although the use of neural networks in this context has a rich history of more than three decades \cite{IdntnCntrlNNNarendra}, recent advances in deep learning \cite{goodfellow2016deep} have led to renewed interest in this topic \cite{watter2015embed, karl2016deep, krishnan2017structured, byravan2017se3, ayed2019learning}. Deep neural networks learn underlying patterns from data and enable generalization beyond the training set by incorporating appropriate inductive bias into the learning approach. To promote representations that are \emph{simple} in some sense, inductive bias \cite{haussler1988quantifying,baxter2000model} often manifests itself via a set of assumptions and guides a learning algorithm to pick one hypothesis over another. The success in predicting an outcome for previously unseen data depends on how well the inductive bias captures the ground reality. Inductive bias can be introduced as the prior in a Bayesian model, or via the choice of computation graphs in a neural network.

Incorporation of physics-based priors into deep learning has been a key focus in the recent times. As these approaches use neural networks to approximate system dynamics, they are more expressive than traditional system identification techniques \cite{soderstrom1988system}. By using a directed graph to capture the causal relationships in a physical system, \cite{sanchez2018graph} introduces a recurrent graph network to infer latent space dynamics in robotic systems. \cite{LutRitPet19} and \cite{gupta2019general} leveraged Lagrangian mechanics to learn the dynamics of kinematic structures from discrete observations. On the other hand, \cite{greydanus2019hamiltonian} and \cite{Zhong2020Symplectic} have utilized Hamiltonian mechanics for learning dynamics from data. However, strict enforcement of the Hamiltonian prior is restrictive for real-life systems which often loses energy in a structured way (e.g. frictional losses in robotic arms, resistive losses in power grids, etc.).

To explicitly encode dissipation as a prior into end-to-end learning, we expand the scope of the Symplectic ODE-Net (SymODEN) architecture \cite{Zhong2020Symplectic} and propose \textit{Dissipative SymODEN}. The underlying dynamics is motivated by the port-Hamiltonian formulation \cite{ortega2002interconnection}, which has a correction term accounting for the prior of dissipation. With this term, Dissipative SymODEN can accommodate the energy losses from various sources of dissipation present in real-life systems. Our results show that inclusion of dissipation into the physics-informed SymODEN architecture improves its prediction accuracy and out-of-sample behavior, while offering insight about relevant physical properties of the system (such as inertia matrix, potential energy, energy dissipation etc.). These insights, in turn, can enable the use of energy-based controllers, such as the method of controlled Lagrangian  \cite{bloch2001controlled} and interconnection \& damping assignment \cite{ortega2002interconnection}, which offer performance guarantees for complex, nonlinear systems.

\textbf{Contribution: }
The main contribution of this work is the introduction of a physics-informed learning architecture called Dissipative SymODEN which encodes a non-conservative physics, i.e. Hamiltonian dynamics with energy dissipation, into deep learning. This provides a means to uncover the dynamics of real-life physical systems whose Hamiltonian aspects have been adapted to external input and energy dissipation. By ensuring that the computation graph is aligned with the underlying physics, we achieve transparency, better predictions with smaller networks, and improved generalization. The architecture of Dissipative SymODEN has also been designed to accommodate angle data in the embedded form. Additionally, we use differentiable ODE solvers to avoid the need for derivative estimation.
\section{The Port-Hamiltonian Dynamics}
Hamiltonian dynamics is often used to systematically describe the dynamics of a physical system in the phase space $(\mathbf{q}, \mathbf{p})$, where $\mathbf{q} = (q_1, q_2, ..., q_n)$ is the generalized coordinate and $\mathbf{p} = (p_1, p_2, ..., p_n)$ is the generalized momentum. In this approach, the key to the dynamics is a scalar function $H(\mathbf{q}, \mathbf{p})$, which is referred to as the Hamiltonian. In almost all physical systems, the Hamiltonian represents the total energy which can be expressed as
\begin{equation}
H(\mathbf{q}, \mathbf{p}) = \frac{1}{2}\mathbf{p}^T \mathbf{M}^{-1}(\mathbf{q}) \mathbf{p} + V(\mathbf{q}),
\label{eqn:H}
\end{equation}
where $\mathbf{M}(\mathbf{q})$ is the symmetric positive definite mass/inertia matrix and $V(\mathbf{q})$ represents the potential energy of the system. The equations of motion are governed by the symplectic gradient~\cite{rowe1980many} of the Hamiltonian, i.e.,
\begin{equation}
    \dot{\mathbf{q}} = \frac{\partial H}{\partial \mathbf{p}} \qquad 
    \dot{\mathbf{p}} = -\frac{\partial H}{\partial \mathbf{q}}.
    \label{Ham_Dyna_Orig}
\end{equation}
Moreover, since $\dot{H} = (\frac{\partial H}{\partial \mathbf{q}})^T \dot{\mathbf{q}} + (\frac{\partial H}{\partial \mathbf{p}} )^T \dot{\mathbf{p}} = 0$, moving along the symplectic gradient conserves the Hamiltonian (i.e. the total energy). However, although the classical Hamiltonian dynamics ensures energy conservation, it fails to model dissipation and external inputs, which often appear in real-life systems. The port-Hamiltonian dynamics generalizes the classical Hamiltonian dynamics by explicitly modelling the total energy, dissipation and external inputs. Motivated by this formulation, we consider the following port-Hamiltonian dynamics in this work:
\begin{equation}
    \label{eqn:H_dyna_damp_force}
    \begin{bmatrix}
        \dot{\mathbf{q}} \\ \dot{\mathbf{p}}
    \end{bmatrix}
    =\bigg(
    \begin{bmatrix}
        \mathbf{0} & \mathbf{I} \\
        -\mathbf{I} & \mathbf{0}
    \end{bmatrix}
    - \mathbf{D}(\mathbf{q}) \bigg)
    \begin{bmatrix}
        \frac{\partial H}{\partial \mathbf{q}}\\
        \frac{\partial H}{\partial \mathbf{p}}
    \end{bmatrix}
    +
    \begin{bmatrix}
        \mathbf{0}\\
        \mathbf{g}(\mathbf{q})
    \end{bmatrix}
    \mathbf{u},
\end{equation}
where the dissipation matrix $\mathbf{D}(\mathbf{q})$ is symmetric positive semi-definite and represents energy dissipation. The external input $\mathbf{u}$ is usually affine and only affects the generalized momenta. The input matrix $\mathbf{g}(\mathbf{q})$ is assumed to have full column rank. As expected, with zero dissipation and zero input, (\ref{eqn:H_dyna_damp_force}) reduces to the classical Hamiltonian dynamics. 
%

\section{Dissipative Symplectic ODE-Net}
\subsection{Training Neural ODE with Constant Forcing}
\label{sec:neural_ode}
%
We focus on the problem of learning an ordinary differential equation (ODE) from observation data. Assume the analytical form of the right hand side (RHS) of an ODE ``$\dot{\mathbf{x}} = \mathbf{f}(\mathbf{x}, \mathbf{u})$" is unknown. An observation data $\mathbf{X} = ((\mathbf{x}_{t_0}, \mathbf{u}_c), ..., (\mathbf{x}_{t_n}, \mathbf{u}_c))$ with a constant input $\mathbf{u}_c$ allows us to approximate $\mathbf{f}(\mathbf{x}, \mathbf{u})$ with a neural net by leveraging
\begin{equation}
    \label{eqn:aug}
    \left.\begin{bmatrix}
        \dot{\mathbf{x}} \\ \dot{\mathbf{u}}
    \end{bmatrix}\right|_{\mathbf{X}}
    =
    \begin{bmatrix}
    \mathbf{f}_{\theta}(\mathbf{x}, \mathbf{u}) \\
        \mathbf{0}
    \end{bmatrix}
    = \tilde{\mathbf{f}}_{\theta}(\mathbf{x}, \mathbf{u}).
\end{equation}
Equation~(\ref{eqn:aug}), by matching the input and output dimensions, enables us to feed it into Neural ODE~\cite{NIPS2018_7892}. With Neural ODE, we make predictions by approximating the RHS of (\ref{eqn:aug}) using a neural network and feed it into an ODE solver
\begin{displaymath}
(\mathbf{\hat{x}}, \mathbf{u}_c)_{t_1, t_2, ..., t_n} = \mathrm{ODESolve}((\mathbf{x}, \mathbf{u}_c)_{t_0}, \tilde{\mathbf{f}}_{\theta}, t_1, t_2, ..., t_n).
\end{displaymath}
We can then construct the loss function $L = \| \mathbf{X} - \mathbf{\hat{X}}\|_2^2$. In practice, we introduce the time horizon $\tau$ as a hyperparameter and predict $\mathbf{x}_{t_{i+1}}, \mathbf{x}_{t_{i+2}}, ..., \mathbf{x}_{t_{i+\tau}}$ from initial condition $\mathbf{x}_{t_i}$, where $i = 0, ..., n-\tau$.
The problem is then how to design the network architecture of $\tilde{\mathbf{f}}_{\theta}$, or equivalently $\mathbf{f}_{\theta}$.

\subsection{Learning from Generalized coordinate and Momentum}
\label{sec:q_and_p}
Suppose we have data consisting of $(\mathbf{q}, \mathbf{p}, \mathbf{u})_{t_0, ..., t_n}$, where $\mathbf{u}$ remains constant in each trajectory. We use four neural nets -- $\mathbf{M}_{\theta_1}^{-1}(\mathbf{q})$, $V_{\theta_2}(\mathbf{q})$, $\mathbf{g}_{\theta_3}(\mathbf{q})$ and $\mathbf{D}_{\theta_4}(\mathbf{q})$ -- as function approximators to represent the inverse of mass matrix, potential energy, the input matrix and the dissipation matrix, respectively. Thus, 
\begin{equation}
    \label{eqn:f_q_p}
    \mathbf{f}_{\theta}(\mathbf{q}, \mathbf{p}, \mathbf{u}) =
    \bigg(
    \begin{bmatrix}
        \mathbf{0} & \mathbf{I} \\
        -\mathbf{I} & \mathbf{0}
    \end{bmatrix}
    - \mathbf{D}_{\theta_4}(\mathbf{q}) \bigg)
    \begin{bmatrix}
        \frac{\partial H_{\theta_1, \theta_2}}{\partial \mathbf{q}}\\
        \frac{\partial H_{\theta_1, \theta_2}}{\partial \mathbf{p}}
    \end{bmatrix}
    +
    \begin{bmatrix}
        \mathbf{0}\\
        \mathbf{g}_{\theta_3}(\mathbf{q})
    \end{bmatrix}
    \mathbf{u}
\end{equation}
where 
\begin{equation}
    H_{\theta_1, \theta_2}(\mathbf{q}, \mathbf{p}) = \frac{1}{2}\mathbf{p}^T \mathbf{M}_{\theta_1}^{-1}(\mathbf{q}) \mathbf{p} + V_{\theta_2}(\mathbf{q}) 
\end{equation}
The partial derivative can be taken care of by automatic differentiation. By putting the designed $\mathbf{f}_{\theta}(\mathbf{q}, \mathbf{p}, \mathbf{u})$ into Neural ODE, we obtain a systematic way of adding the prior knowledge of a structured dynamics into end-to-end learning.

\subsection{Learning from Embedded Angle Data}
\label{sec:embed_data}
Often, especially in robotics, the state variables involve angles residing in the interval $[-\pi, \pi)$. In other words, each angle lies on the manifold $\mathbb{S}^1$. However, generalized coordinates are typically assumed to lie on $\mathbb{R}^n$. To bridge this gap, we use an angle-aware design~\cite{Zhong2020Symplectic} and assume that the generalized coordinates are angles available as $(\mathbf{x}_1(\mathbf{q}), \mathbf{x}_2(\mathbf{q}), \mathbf{x}_3(\dot{\mathbf{q}}), \mathbf{u})_{t_0, ..., t_n} = (\cos \mathbf{q}, \sin \mathbf{q}, \dot{\mathbf{q}}, \mathbf{u})_{t_0, ..., t_n}$. Then, similar to \cite{Zhong2020Symplectic}, we aim to learn a structured dynamics (\ref{eqn:H_dyna_damp_force}) expressed in terms of $\mathbf{x}_1$, $\mathbf{x}_2$ and $\mathbf{x}_3$. As $\mathbf{p} = \mathbf{M}(\mathbf{x}_1, \mathbf{x}_2)\dot{\mathbf{q}}$, we can express this dynamics as
\begin{align}
    \label{eqn:angular_data}
    \dot{\mathbf{x}}_1 &= -\sin \mathbf{q} \circ \dot{\mathbf{q}} = - \mathbf{x}_2  \circ \dot{\mathbf{q}} \nonumber\\ 
    \dot{\mathbf{x}}_2 &= \cos \mathbf{q} \circ \dot{\mathbf{q}} = \mathbf{x}_1 \circ \dot{\mathbf{q}} \\
    \dot{\mathbf{x}}_3 &= \frac{\mathrm{d}}{\mathrm{d}t}(\mathbf{M}^{-1}(\mathbf{x}_1, \mathbf{x}_2)\mathbf{p}) = \frac{\mathrm{d}}{\mathrm{d}t}(\mathbf{M}^{-1}(\mathbf{x}_1, \mathbf{x}_2)) \mathbf{p} + \mathbf{M}^{-1}(\mathbf{x}_1, \mathbf{x}_2) \dot{\mathbf{p}}, \nonumber
\end{align}
where ``$\circ$" represents the element-wise product. We assume $\mathbf{q}$ and $\mathbf{p}$ evolve with the structured dynamics Equation~(\ref{eqn:H_dyna_damp_force}) and substitute Equation~(\ref{eqn:H_dyna_damp_force}) in to the RHS of Equation~(\ref{eqn:angular_data}). Similar to our approach in Sec~\ref{sec:q_and_p}, we use four neural nets to express the RHS of Equation~(\ref{eqn:angular_data}) as $\mathbf{f}_{\theta}(\mathbf{x}_1, \mathbf{x}_2, \mathbf{x}_3, \mathbf{u})$. Thus, it can be fed into Equation \ref{eqn:aug} and the Neural ODE.

\subsection{Learning on Hybrid Spaces $\R^n\times\mathbb{T}^m$}
\label{sec:hybrid}
In most of physical systems, both translational coordinates and rotational coordinates coexist. In other words, the generalized coordinates lie on $\R^n\times\mathbb{T}^m$, where $\mathbb{T}^m$ denotes the $m$-torus. Here we put together the architecture of the previous two subsections. We assume the generalized coordinates are  $\mathbf{q} =  (\mathbf{r}, \pmb{\phi}) \in \R^n\times\mathbb{T}^m$ and the data comes in the form of $(\mathbf{x}_1, \mathbf{x}_2, \mathbf{x}_3, \mathbf{x}_4, \mathbf{x}_5, \mathbf{u})_{t_0, ..., t_n} = (\mathbf{r}, \cos \pmb{\phi}, \sin \pmb{\phi}, \dot{\mathbf{r}}, \dot{\pmb{\phi}}, \mathbf{u})_{t_0, ..., t_n}$. We use four neural nets -- $\mathbf{M}_{\theta_1}^{-1}(\mathbf{x}_1, \mathbf{x}_2, \mathbf{x}_3)$, $V_{\theta_2}(\mathbf{x}_1, \mathbf{x}_2, \mathbf{x}_3)$, $\mathbf{g}_{\theta_3}(\mathbf{x}_1, \mathbf{x}_2, \mathbf{x}_3)$ and $\mathbf{D}_{\theta_4}(\mathbf{x}_1, \mathbf{x}_2, \mathbf{x}_3)$ -- as function approximators. Then the dynamics is given by
$$[\dot{\mathbf{x}}_1, \dot{\mathbf{x}}_2, \dot{\mathbf{x}}_3, \dot{\mathbf{x}}_4, \dot{\mathbf{x}}_5]^T = \mathbf{f}_{\theta}(\mathbf{x}_1, \mathbf{x}_2, \mathbf{x}_3, \mathbf{x}_4, \mathbf{x}_5, \mathbf{u}) $$

\subsection{The Dissipation Matrix and the Mass matrix}
%
As the dissipation matrix models energy dissipation such as friction and resistance, it is positive semi-definite. We impose this constraint in the network architecture by $\mathbf{D}_{\theta_4} = \mathbf{L}_{\theta_4}\mathbf{L}_{\theta_4}^T$, where $\mathbf{L}_{\theta_4}$ is a lower-triangular matrix. In real physical systems, both the mass matrix $\mathbf{M}$ and its inverse are positive definite. Similarly, semi-definiteness is constraint by $\mathbf{M}_{\theta_1}^{-1} = \mathbf{L}_{\theta_1}\mathbf{L}_{\theta_1}^T$, where $\mathbf{L}_{\theta_1}$ is a lower-triangular matrix. The positive definiteness is ensured by adding a small constant $\epsilon$ to the diagonal elements of $\mathbf{M}^{-1}_{\theta_1}$. It not only makes $\mathbf{M}_{\theta_1}$ invertible, but also stabilizes training.

\section{Experiments}

\subsection{Experimental Setup}
%
We use the following four tasks to evaluate the performance of Dissipative SymODEN architecture -- (i) Task 1: a pendulum with generalized coordinate and momentum data; (ii) Task 2: a pendulum with embedded angle data; (iii) Task 3: a CartPole system; and (iv) Task 4: an Acrobot. 

\textbf{Model Variants: }Besides the Dissipative SymODEN model derived above, we consider a variant, called \textit{Unstructured (Unstr.) Dissipative SymODEN}, which approximates the Hamiltonian by a fully connected neural net $H_{\theta_1, \theta_2}$. We also consider the original \textit{SymODEN} \cite{Zhong2020Symplectic} as a model variant.

\textbf{Baseline Models: }We set up baseline models for all four experiments. For the pendulum with generalized coordinate and momentum data, the \textit{naive baseline} model approximates (\ref{eqn:f_q_p}) -- $\mathbf{f}_{\theta}(\mathbf{x}, \mathbf{u})$ -- by a fully connected neural net. For all the other experiments, which involves embedded angle data, we set up two different baseline models: \textit{naive baseline} approximates $\mathbf{f}_{\theta}(\mathbf{x}, \mathbf{u})$ by a fully connected neural net. Also, we set up the \textit{geometric baseline} model which approximates $\dot{\mathbf{q}}$ and $\dot{\mathbf{p}}$ with a fully connected neural net.

\textbf{Data Generation: }For all tasks, we randomly generated initial conditions of states and subsequently combined them with 5 different constant control inputs, i.e., $u = -2.0, -1.0, 0.0, 1.0, 2.0$, to produce the initial conditions and input required for simulation. The simulators integrate the corresponding dynamics for 20 time steps to generate trajectory data which is then used to construct the training set and test set.

\textbf{Model training: }In all the tasks, we train our model using Adam optimizer \cite{2014arXiv1412.6980K} with 1000 epochs. We set a time horizon $\tau=3$, and choose ``RK4" as the numerical integration scheme in Neural ODE. We logged the \textit{train error}, \textit{test error} and \textit{prediction (pred.) error} per trajectory for all the tasks. Prediction error per trajectory is calculated by using the same initial state condition in the training set with a constant control of $u=0.0$, integrating 40 time steps forward.

\subsection{Task 1: Pendulum with Generalized Coordinate and Momentum Data}
\label{sec:task1}
\begin{wrapfigure}[15]{r}{0.60\textwidth}
\centering
\vspace{-0.5em}
\includegraphics[width=0.59\textwidth]{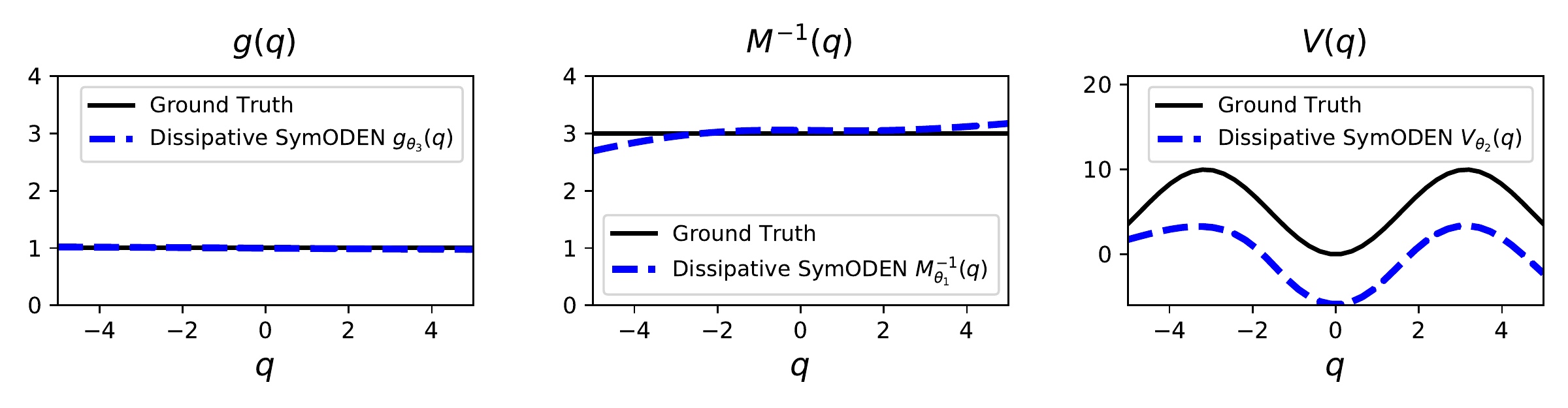}
\\
\vspace{-0.55em}
\includegraphics[width=0.6\textwidth]{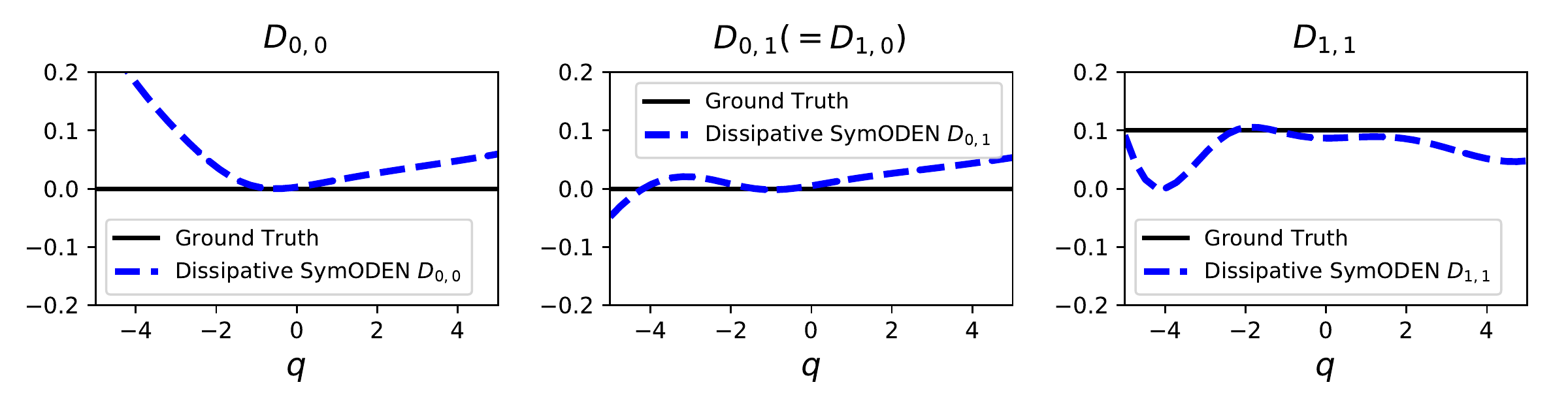}
\vspace{-2.95em}
\caption{\small Learned functions in Task 1 (Pendulum). }
\vspace{-22pt}
\label{fig:pend}
\end{wrapfigure}
In this task, we use the model described in Section \ref{sec:q_and_p} and present the predicted trajectories of the learned models as well as the learned functions of Dissipative SymODEN. The underlying dynamics is given by
\begin{align}
    \dot{q} = 3p, \qquad
    \dot{p} = -5 \sin q - 0.3p+ u,
    \label{eqn:pend}
\end{align}
with the Hamiltonian $H(q, p) = 1.5 p^2 + 5(1-\cos q)$. In other words $M^{-1}(q)=3$, $V(q)=5(1-\cos q)$, $g(q)=1$ and $\mathbf{D}_{\theta_4}(q)= [0,0;0,0.1]$. Figure~\ref{fig:pend} shows that the learned $g_{\theta_3}(q)$ and $M^{-1}_{\theta_1}(q)$ matches the ground truth pretty well. Also, $V_{\theta_2}(q)$ differs from the ground truth by an almost constant margin which is expected since only the derivative of $V_{\theta_2}(q)$ impacts the dynamics. The learned dissipation matrix $\mathbf{D}_{\theta_4}(q)$ does not match the ground truth. We address this issue in the next subsection.
In Table \ref{tab:errors}, Naive Baseline's prediction error is the lowest because predicted trajectories reach the origin faster than the ground truth.
%
\begin{figure}[htbp]
    \centering
    \includegraphics[width=0.8\textwidth]{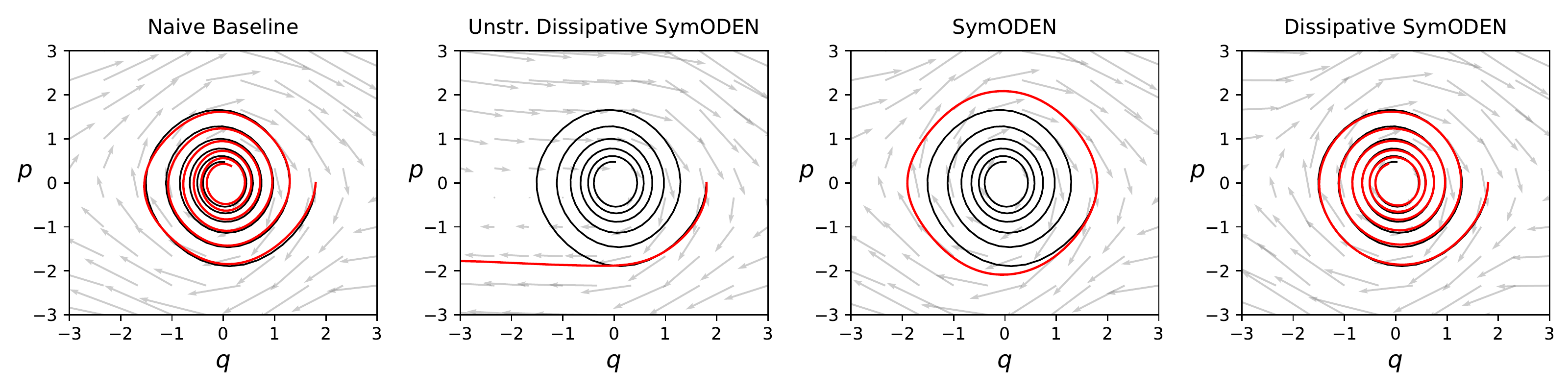}
    \vspace{-15pt}
    \caption{\small Learned trajectories of different models. Red and black lines represent the learned and ground truth trajectories, respectively and the gray arrows show the vector fields learned by each model. \textit{Dissipative SymODEN} learns a more accurate vector field than the \textit{naive baseline} model. Moreover, it appears that whereas \textit{SymODEN} learns an energy-conserved vector field slightly different from the ground truth, \textit{Unstructured Dissipative SymODEN} learns it completely wrong. }
    \label{fig:damp-pend-traj}
\end{figure}

\subsection{Task 2: Pendulum with Embedded Data}
\begin{wrapfigure}[13]{r}{0.60\textwidth}
    \centering
    \vspace{-0.95em}
    \includegraphics[width=0.59\textwidth]{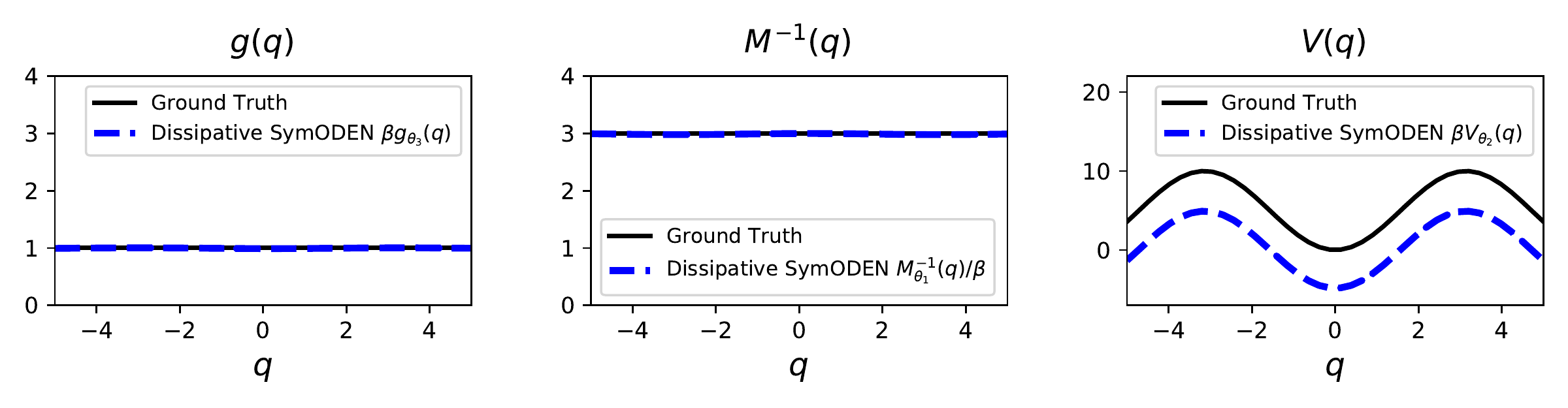}\\
    \vspace{-0.55em}
    \includegraphics[width=0.6\textwidth]{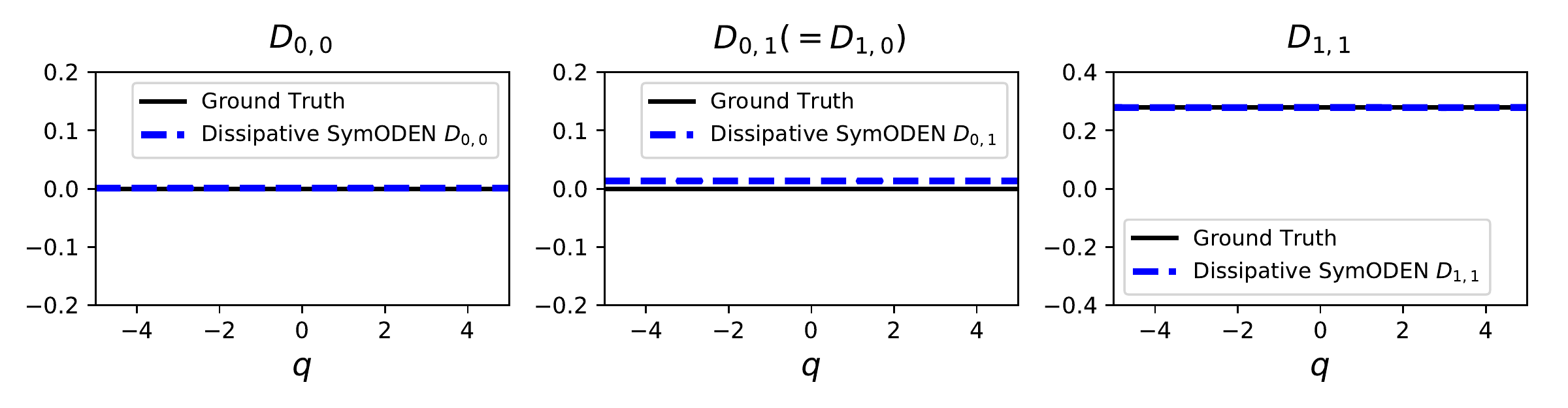}
    \vspace{-2.95em}
    \caption{\small Learned functions in Task 2 (Pendulum with embedded data).}
    \vspace{-22pt}
    \label{fig:embed}
\end{wrapfigure}
In this task, the dynamics is the same as Equation~(\ref{eqn:pend}) but the training data are generated by the OpenAI Gym simulator, i.e. we use embedded angle data and assume we only have access to $\dot{q}$ instead of $p$. We use the model described in Section \ref{sec:embed_data} to learn the structured dynamics. Without true $p$ data, the learned function matches the ground truth with a scaling $\beta$, as shown in Figure \ref{fig:embed}. Please refer to \cite{Zhong2020Symplectic} for explanation of the scaling. 
In this example, with the scaling $\beta=0.357$, the learned functions match the ground truth. With the angle-aware design, we learned the dissipation matrix much better than the previous subsection. 

\subsection{Results}
In Table \ref{tab:errors}, we show the train, test and prediction errors for all four tasks. Dissipative SymODEN performs the best in all three metrics. As SymODEN does not allow dissipation, it does not perform well in these tasks. Since Unstructured Dissipative SymODEN architecture has trouble learning a good vector field, it performs the worst in all the tasks except Task 2. In conclusion, Dissipative SymODEN achieves higher accuracy with less model parameters. Moreover, the learned model reveals physical aspects of the system, which can be leveraged by energy-based controllers.
{\small
\begin{table}[h!]
\caption{\small Train, Test and Prediction Errors of Four Tasks}
\label{tab:errors}
\centerline{
\begin{tabular}{c l l l l l l} 
    \toprule[1pt]
    \textbf{Task} & & \centering\shortstack{Naive \\ Baseline} &  \centering\shortstack{Geometric \\ Baseline} & \centering\shortstack{UnStr. Dissipative \\ SymODEN} &  SymODEN & \shortstack{Dissipative \\ SymODEN}\\
    \midrule[1pt]
    \textbf{1} & \textit{\#Parameters} &  $0.36$M & N/A & $0.22$M & $0.13$M & $0.15$M \\
    \cmidrule[0.5pt](lr){2-7}
    & Train error & $26.38 \pm 38.00$ & N/A & $34.80 \pm 68.53$ & $4.47 \pm 6.40$ & $0.88 \pm 1.41$ \\
    & Test error  & $35.03 \pm 49.89$ & N/A & $49.44 \pm 81.31$ & $7.52 \pm 10.13$ & $1.25 \pm 1.81$ \\
    & Pred. error & $32.544 \pm 36.203$ & N/A & $219.36 \pm 296.86$ & $96.50 \pm 99.56$ & $34.03 \pm 47.83$ \\
    \midrule[1pt]
    \textbf{2} & \textit{\#Parameters} &  $0.65$M & $0.46$M & $0.41
    $M & $0.14$M & $0.16$M\\
    \cmidrule[0.5pt](lr){2-7}
    & Train error & $2.02 \pm 4.41$ & $0.42 \pm 1.16$ & $1.90 \pm 3.85$ & $2.37 \pm 2.71$ & $0.15 \pm 0.27$ \\
    & Test error  & $2.01 \pm 4.99$ & $0.33 \pm 1.22$ & $1.61 \pm 3.36$ & $2.67 \pm 2.83$ & $0.13 \pm 0.25$ \\
    & Pred. error & $40.18 \pm 78.10$ & $0.81 \pm 0.68$ & $7.04 \pm 13.65$ & $72.78 \pm 90.42$ & $1.04 \pm 1.3$\\
    \midrule[1pt]
    \textbf{3} & \textit{\#Parameters} &  $1.01$M & $0.82$M & $0.69$M & $0.51$M & $0.53$M\\
    \cmidrule[0.5pt](lr){2-7}
    & Train error & $12.92 \pm 15.58$ & $0.48 \pm 0.50$ & $12.09 \pm 18.38$ & $3.33 \pm 3.85$ & $0.88 \pm 0.89$ \\
    & Test error  & $20.07 \pm 26.42$ & $1.34 \pm 3.19$ & $19.87 \pm 23.16$ & $3.80 \pm 3.71$ & $1.37 \pm 1.30$ \\
    & Pred. error & $268.24 \pm 204.15$ & $60.12 \pm 96.18$ & $366.38 \pm 405.45$ & $30.21 \pm 34.33$ & $8.32 \pm 7.81$\\
    \midrule[1pt]
    \textbf{4} & \textit{\#Parameters} &  $1.46$M & $0.97$M & $0.80$M & $0.51$M & $0.53$M\\
    \cmidrule[0.5pt](lr){2-7}
    & Train error & $1.76 \pm 2.26$ & $1.90 \pm 2.82$ & $77.56 \pm 111.50$ & $2.92 \pm 2.58$ & $0.47 \pm 0.64$ \\
    & Test error  & $5.12 \pm 9.14$ & $4.87 \pm 7.42$ & $122.70 \pm 190.90$ & $5.27 \pm 6.55$ & $0.81 \pm 1.10$ \\
    & Pred. error & $36.65 \pm 77.16$ & $44.26 \pm 95.70$ & $590.77 \pm 807.88$ & $68.26 \pm 103.46$ & $12.72 \pm 32.12$\\
    \bottomrule[1pt]
\end{tabular}}
\end{table}
}
\subsubsection*{Acknowledgments}
This research was inspired by the ideas and plans articulated by N. E. Leonard and A. Majumdar, Princeton University, in their ONR grant \#N00014-18-1-2873.  The research was primarily carried out during Y. D. Zhong's internship at Siemens Corporation, Corporate Technology. Pre- and post-internship, Y. D. Zhong's work was supported by ONR grant \#N00014-18-1-2873.
\bibliography{Dissipative_SymODEN_Refs}

\begin{thebibliography}{10}
\providecommand{\url}[1]{#1}
\csname url@samestyle\endcsname
\providecommand{\newblock}{\relax}
\providecommand{\bibinfo}[2]{#2}
\providecommand{\BIBentrySTDinterwordspacing}{\spaceskip=0pt\relax}
\providecommand{\BIBentryALTinterwordstretchfactor}{4}
\providecommand{\BIBentryALTinterwordspacing}{\spaceskip=\fontdimen2\font plus
\BIBentryALTinterwordstretchfactor\fontdimen3\font minus
  \fontdimen4\font\relax}
\providecommand{\BIBforeignlanguage}[2]{{%
\expandafter\ifx\csname l@#1\endcsname\relax
\typeout{** WARNING: IEEEtran.bst: No hyphenation pattern has been}%
\typeout{** loaded for the language `#1'. Using the pattern for}%
\typeout{** the default language instead.}%
\else
\language=\csname l@#1\endcsname
\fi
#2}}
\providecommand{\BIBdecl}{\relax}
\BIBdecl

\bibitem{lillicrap2015continuous}
T.~P. Lillicrap, J.~J. Hunt, A.~Pritzel, N.~Heess, T.~Erez, Y.~Tassa,
  D.~Silver, and D.~Wierstra, ``Continuous control with deep reinforcement
  learning,'' \emph{arXiv:1509.02971}, 2015.

\bibitem{Wei:2017:DRL:3061639.3062224}
T.~Wei, Y.~Wang, and Q.~Zhu, ``{Deep Reinforcement Learning for Building HVAC
  Control},'' in \emph{Proceedings of the 54th Annual Design Automation
  Conference (DAC)}, 2017, pp. 22:1--22:6.

\bibitem{IdntnCntrlNNNarendra}
K.~S. {Narendra} and K.~{Parthasarathy}, ``Identification and control of
  dynamical systems using neural networks,'' \emph{IEEE Transactions on Neural
  Networks}, vol.~1, no.~1, pp. 4--27, 1990.

\bibitem{goodfellow2016deep}
I.~Goodfellow, A.~Courville, and Y.~Bengio, \emph{Deep learning}.\hskip 1em
  plus 0.5em minus 0.4em\relax MIT Press, 2016, vol.~1.

\bibitem{watter2015embed}
M.~Watter, J.~Springenberg, J.~Boedecker, and M.~Riedmiller, ``Embed to
  control: A locally linear latent dynamics model for control from raw
  images,'' in \emph{{Advances in Neural Information Processing 29}}, 2015, pp.
  2746--2754.

\bibitem{karl2016deep}
M.~Karl, M.~Soelch, J.~Bayer, and P.~van~der Smagt, ``Deep variational bayes
  filters: Unsupervised learning of state space models from raw data,''
  \emph{arXiv:1605.06432}, 2016.

\bibitem{krishnan2017structured}
R.~G. Krishnan, U.~Shalit, and D.~Sontag, ``Structured inference networks for
  nonlinear state space models,'' in \emph{Thirty-First AAAI Conference on
  Artificial Intelligence}, 2017.

\bibitem{byravan2017se3}
A.~Byravan and D.~Fox, ``Se3-nets: Learning rigid body motion using deep neural
  networks,'' in \emph{2017 IEEE International Conference on Robotics and
  Automation (ICRA)}.\hskip 1em plus 0.5em minus 0.4em\relax IEEE, 2017, pp.
  173--180.

\bibitem{ayed2019learning}
I.~Ayed, E.~de~B{\'e}zenac, A.~Pajot, J.~Brajard, and P.~Gallinari, ``Learning
  dynamical systems from partial observations,'' \emph{arXiv:1902.11136}, 2019.

\bibitem{haussler1988quantifying}
D.~Haussler, ``{Quantifying inductive bias: AI learning algorithms and
  Valiant's learning framework},'' \emph{Artificial Intelligence}, vol.~36,
  no.~2, pp. 177--221, 1988.

\bibitem{baxter2000model}
J.~Baxter, ``A model of inductive bias learning,'' \emph{Journal of Artificial
  Intelligence Research}, vol.~12, pp. 149--198, 2000.

\bibitem{soderstrom1988system}
T.~S{\"o}derstr{\"o}m and P.~Stoica, \emph{System identification}.\hskip 1em
  plus 0.5em minus 0.4em\relax Prentice-Hall, Inc., 1988.

\bibitem{sanchez2018graph}
A.~Sanchez-Gonzalez, N.~Heess, J.~T. Springenberg, J.~Merel, M.~Riedmiller,
  R.~Hadsell, and P.~Battaglia, ``Graph networks as learnable physics engines
  for inference and control,'' in \emph{International Conference on Machine
  Learning (ICML)}, 2018, pp. 4467--4476.

\bibitem{LutRitPet19}
M.~Lutter, C.~Ritter, and J.~Peters, ``Deep lagrangian networks: Using physics
  as model prior for deep learning,'' in \emph{7th International Conference on
  Learning Representations (ICLR)}, 2019.

\bibitem{gupta2019general}
J.~K. Gupta, K.~Menda, Z.~Manchester, and M.~J. Kochenderfer, ``A general
  framework for structured learning of mechanical systems,''
  \emph{arXiv:1902.08705}, 2019.

\bibitem{greydanus2019hamiltonian}
S.~{Greydanus}, M.~{Dzamba}, and J.~{Yosinski}, ``{Hamiltonian Neural
  Networks},'' \emph{arXiv:1906.01563}, 2019.

\bibitem{Zhong2020Symplectic}
Y.~D. Zhong, B.~Dey, and A.~Chakraborty, ``{Symplectic ODE-Net: Learning
  Hamiltonian Dynamics with Control},'' in \emph{International Conference on
  Learning Representations (ICLR)}, 2020.

\bibitem{ortega2002interconnection}
R.~Ortega, A.~J. Van Der~Schaft, B.~Maschke, and G.~Escobar, ``Interconnection
  and damping assignment passivity-based control of port-controlled hamiltonian
  systems,'' \emph{Automatica}, vol.~38, no.~4, pp. 585--596, 2002.

\bibitem{bloch2001controlled}
A.~M. Bloch, N.~E. Leonard, and J.~E. Marsden, ``Controlled lagrangians and the
  stabilization of euler--poincar{\'e} mechanical systems,''
  \emph{International Journal of Robust and Nonlinear Control}, vol.~11, no.~3,
  pp. 191--214, 2001.

\bibitem{rowe1980many}
D.~J. Rowe, A.~Ryman, and G.~Rosensteel, ``Many-body quantum mechanics as a
  symplectic dynamical system,'' \emph{Physical Review A}, vol.~22, no.~6, p.
  2362, 1980.

\bibitem{NIPS2018_7892}
T.~Q. Chen, Y.~Rubanova, J.~Bettencourt, and D.~K. Duvenaud, ``Neural ordinary
  differential equations,'' in \emph{{Advances in Neural Information Processing
  Systems 31}}, 2018, pp. 6571--6583.

\bibitem{2014arXiv1412.6980K}
D.~P. {Kingma} and J.~{Ba}, ``{Adam: A Method for Stochastic Optimization},''
  \emph{arXiv:1412.6980}, 2014.

\end{thebibliography}
\bibliographystyle{IEEEtran}
\end{document}